# RRLFSOR: An Efficient Self-Supervised Learning Strategy of Graph Convolutional Networks

Feng Sun[§], Ajith Kumar V[§], Guanci Yang[*], Qikui Zhu[*], Yiyun Zhang, Ansi Zhang, Dhruv Makwana

Feng Sun
Postgraduate, Assistant Experimentalist
Experimental Teaching Center for Liberal Arts, Zhejiang Normal University, Jinhua, Zhejiang, China 321004
Tel: 86-131 5799 3079
Email: luckysunfeng@126.com

Ajith Kumar V
Postgraduate, Senior Engineer
School of AI, Bangalore, India 560002
Tel: 91-91 5065 9274
Email: inocajith21.5@gmail.com

Guanci Yang
Ph.D, Professor
Key Laboratory of Advanced Manufacturing Technology of Ministry of Education, Guizhou University, Guiyang, Guizhou, China 550025
Tel: 86-0851-8437-007
Email: gcyang@gzu.edu.cn

Qikui Zhu
PhD, Lecturer
The Department of Medical Imaging, Western University, London, Canada W1k 6NG
Tel: +86-156 2347 1424
Email: QikuiZhu@163.com

Yiyun Zhang
Bachelor Student
Engineering College, Zhejiang Normal University, Jinhua, Zhejiang, China 321004
Tel: 86-151 6587 8848
Email: zhangyuni@zjnu.edu.cn

Ansi Zhang
PhD, Lecturer
School of Mechanical Engineering, Guizhou University, Guiyang, Guizhou, China 550025
Tel: 86-188 8499 1757
Email: zhangas@gzu.edu.cn

Dhruv Makwana

Bachelor, Senior Engineer

Computer Science College, Gujarat Technological University, Bangalore, India 560002

Tel: 91-9904278934

Email: dmakwana503@gmail.com

§ These two authors contributed equally

* Corresponding authors

**ABSTRACT**

Graph Convolutional Networks (GCNs) are widely used in many applications yet still need large amounts of labelled data for training. Besides, the adjacency matrix of GCNs is stable, which makes the data processing strategy cannot efficiently adjust the quantity of training data from the built graph structures. To further improve the performance and the self-learning ability of GCNs, in this paper, we propose an efficient self-supervised learning strategy of GCNs, named randomly removed links with a fixed step at one region (RRLFSOR). RRLFSOR can be regarded as a new data augmenter to improve over-smoothing. RRLFSOR is examined on two efficient and representative GCN models with three public citation network datasets – Cora, PubMed, and Citeseer. Experiments on transductive link prediction tasks show that our strategy outperforms the baseline models consistently by up to 21.34% in terms of accuracy on three benchmark datasets.

1. Introduction

Graph Convolutional Networks (GCNs) extend Convolutional Neural Networks (CNNs), processing complex graph structure data. It has been widely applied to action recognition, semantic segmentation[1], attribute recognition[2], point cloud classification and image classification. For example, Permutoheadral-GCN[3] was proposed, a global attention mechanism, and its node can automatically attend or not and whether aggregate from other nodes. Typically, GCNs acquire information from their neighbourhoods. To exploit the information from long-range nodes inside the graph, GAT-POS[4] was proposed, which improved the ability to obtain positional information of the nodes of GAT. Geom-GCN[5] was proposed to exploit the structural information of nodes between neighbourhoods and acquire information from long-range by consisting of bi-level aggregation, structural neighbourhood, and node embedding.

Additionally, WGCN[6] was proposed, which could acquire information from long-range nodes and the local topologies by utilizing weighted structural features. However, not all neighbour nodes are significant for the target node, and it's hard to capture long-range non-local information from other nodes. To solve the above problem, Roy et al.[7] defined two types of neighbourhoods: local neighbourhoods and non-local neighbourhoods. Local neighbourhoods generated a community between nodes that could share mutual information with its neighbours. Non-local neighbourhoods built a node clustering to make the distant nodes in the same cluster.

Indeed, another two challenges for GCNs still need to be solved[8]. The first one is that graph convolutional networks need to go deeper. That is to say, and when the number of layers is more than 3, current GCNs will suffer from high computational costs and high memory usage. To tackle this problem, L-GCN[9] was proposed by designing an efficient layer-wise training framework. The second one is most GCNs are static and are not suitable for scenes with dynamic features. The EvolveGCN[10] model was utilized to solve the problem of fixed and less applicable to node sets' continual changes inside GCNs. To solve the existing GNNs are improper in the dynamic graphs, the TM-GCN[11] was introduced, which used a tensor algebra mechanism. TM-GCN is a consistent framework which is for message passing neural networks. Manessi et al. [12] developed a model that could express the changing structure of the graph by combining GCNs and LSTM, which aimed to learn the long short-term dependences with the graph. In addition, preserving high-order proximity is very important for dynamic networks. DHPE[13] was proposed, which aimed to solve most network neglect the changing feature of real-world applications which cannot preserve the high-order proximity very well.

Self-Supervised Learning (SSL) is widely used in natural language processing, image noise reduction, and computer vision for overcoming the problem of the limited labelled dataset. But existing methods only focus on preserving the local similarity structure and ignoring the global structure of all datasets. To address this issue, Xu et al. [14] proposed a GraphLoG for self-supervised learning. FedGL[15] was offered, which could acquire global patterns and protect privacy by digging into the global self-supervised information. To learn graph presentation with a few supervised labelled nodes that are difficult to solve, the M3S[16] training algorithm was proposed. Facing the traditional graph are simple, and they have poor generalization because of manual designs，InfoGraph [17] was realized to learn the graph representations. You et al. [18] proved that it could improve the performance of GCNs when adopting appropriate self-supervised learning tasks. So we apply link prediction as the task in the model to enhance the performance.

Despite the fruitful progress, yet, there are some problems that have to solve also. On the one hand, GNN and GCNs require a lot of labelled data in network training. On the other hand, over-fitting and over-smoothing are two main limitations of GCNs for link prediction especially for over-smoothing. To address these limitations, we propose a self-supervised learning strategy inspired by [19]. We propose an efficient self-supervised learning strategy of GCNs, named randomly removing links with a fixed step at one region. We use this strategy to modify the input data. Then input the data into the model. Finally, we use link prediction as our task to verify our strategy. This strategy can be regarded as a data augmenter. Removing certain links with steps is making node connections more sparse, and avoids over-smoothing.

In all, our contributions can be summarized as the following.

**1.** We propose an efficient self-supervised learning strategy of GCNs, named randomly removing links with a fixed step at one region (**RRLFSOR**). RRLFSOR is not constrained by labelled data and obtains feature data directly from graph structure data.

**2.** To the best of our knowledge**, RRLFSOR** is the first strategy that could increase the accuracy of GCN by 21.34% in the Citeseer dataset. To verify the efficiency of our strategy, we also apply our strategy to L-GCN to demonstrate our strategy, which is a successful and representative variant of GCN. L-GCN is a layer-wise GCN variant and not affected by the layer number. L-GCN was published in CVPR2020. Meanwhile, **RRLFSOR** increases 8.39% of the accuracy than L-GCN in the PubMed dataset.

**3.** Extensive experiments are performed on three public citation network datasets to show significantly improve the performance of the efficient and representative GCNs models. More importantly, the strategy we proposed is universal and portable. In a word, the strategy we proposed could be applied to any GCN model in a plug-and-play way.

The remainder of this paper is organized as follows. Section 2 introduces the related work. Section 3 reviews some preliminaries. Section 4 describes the proposed RRLFSOR in detail. Section 5 sincerely presents the experiments and analysis. Lastly, we present the conclusions in Section 6.

## 2. Related work

### 2.1 Graph convolutional networks

Recently, Graph Convolutional Networks(GCNs) has become one of the most technology in many applications, including image classification and action recognition. GCNs is a good technology to solve tasks with non-Euclidean data. For example, link prediction[20-21] and node classification[22-24]. Rong et al. [25] proposed DropEdge to solve the over-smoothing and over-fitting of GCNs. DropEdge only explored the performance in the whole graph and ignored the neighbourhood of nodes. GCN[26] was realized by choosing the convolutional model via a localized first-order approximation of spectral graph convolutions. GAT[27] was introduced to leverage masked self-attentional layers. Yang et al. [28] proposed DGCs, a method for calculating over-fitting through dropout in the multi-layer graph convolutional networks. Zhu et al. [29] utilized two self-supervised learning(SSL) strategies to capture information from the graph structure data. The first one is randomly removing links(RRL). The second one is randomly covering features(RCF). These two strategies are used to modify the input data. Then to input the data into the SSL model. Finally, applied the SSL model in GCN and GAT. To deliver the best performance for semi-supervised learning tasks, Qin et al. [30] defined Graph convolutional network with estimated labels(E-GCN). What's more, E-GCN could solve the unknown labels evaluation of the graph convolution. To improve the performance of semi-supervised learning on GCNs, N-GCN[31] was proposed. N-GCN trains multiple instances of GCNs and optimizes the classification objective by combining the instance outputs. In addition, Cross-GCN[32] was presented to improve the robustness and performance of GCNs by crossing features. It's worth noting that Cross-GCN explicitly models the arbitrary-order cross features.

Moreover, adversarial training(AT) is an effective technique to enhance the robustness of neural networks. Some works apply the AT in GCNs directly and improve their performance and robustness. Sun et al. [33] utilized Virtual adversarial training(VAT), which was based on both unlabeled and labelled data to improve the generalization performance of GCNs. Feng et al. [34] applied adversarial training on the graph to enhance the generalization and robustness of models. To this end, he presented Graph adversarial training(GraphAT). GraphAT is a dynamic scheme based on the graph structure. Hu et al. [35] proposed directional graph adversarial training(DGAT), a model to apply the graph structure in the adversarial process and identify the impact of perturbations automatically from its neighbour nodes.

### 2.2 Self-supervised learning

Since deep learning relies on a large amount of labelled data, sometimes the labelled data for training models is limited, which causes these models to be over-fitting. Self-supervised learning(SSL) has gained a lot of attention to solve this problem. For example, Zeng et al. [36] introduced two methods based on contrastive self-supervised learning(CSSL). The first method is to utilize CSSL to pretrain the encoders of the graph on unlabeled graphs without being based on labels, then adjust the pre-trained encoders on the labelled graphs. The second method is to present a regularizer based on CSSL, which is to alleviate the unsupervised CSSL task and the supervised classification task simultaneously. Meanwhile, SelfSAGCN [37] was defined to address the poor performance when the labelled data are scarce. SelfSAGCN consists of two methods. The first method is Semantic Alignment which is used to align node features acquired from different aspects. The second method is Identity Aggregation which is utilized to extract semantic features from the labelled nodes. Indeed, they could alleviate the over-smoothing problem.

Recently, most existing studies use SSL to capture subgraph representations among graphs. For example, Sun et al. [38] introduced a self-supervised mutual information method to enhance subgraph embedding of the global graph structural properties, SUGAR. SUGAR captures the subgraph representations among graphs, which are explicit of both global and local properties. Besides, SL-GAD [39] was proposed to construct different contextual subgraphs depending on the target node of the entire graph. SL-GAD generated multi-view contrastive learning and attributes regression for anomaly detection. The multi-view contrastive learning method can capture richer structure information from the subgraphs.

Moreover, SSL has been applied in molecular property prediction recently. For instance, Zhao et al. [40] utilized CSGNN which was to predict molecular interactions. CSGNN solves existing models that heavily rely on structures or features involving molecules. CSGNN acquires high-order dependency by injecting the mix-hop neighbourhood aggregator into GNN. Recently, most existing self-supervised pre-training methods for GNNs only focus on graph-level or node-level tasks. To bridge this gap, MGSSL[41] was proposed. MGSSL is a method of self-supervised motif generation framework for GNNs.

### 2.3 Data augmentation

Data augmentation has emerged as a successful method for deep learning structures. It changes the input data directly instead of modifying the model architecture, which could improve the effectiveness significantly. Recently, data augmentation are widely used in the image domain and graph domain. In the image domain, there are some data augmentation techniques. Such as colour space, flipping, cropping, random erasing, translation and noise injection [42]. Zhong et al. [43] first proposed a random erasing technique. As the variants of random erasing, Pathak et al. [19] proposed two methods to deal with the input data. The first one is the random block, which means randomly removing some block images. The second one is the random region, which means randomly removing one region images. After generating images, then input them into the model to inpainting the images. This paper is inspired by these methods. Moreno-Barea et al. [44] verified noise injection by adjusting the degree of noise for individual input variables. Zhou et al. [45] proposed four augmentation methods on graphs to overcome undergeneralization and over-fitting of the limitation of scale in the benchmark dataset for graph classification. The four methods are motif-random mapping, random mapping, vertex-similarity mapping and motif-similarity mapping. Meanwhile, contrastive learning has emerged as a successful technique for graph augmentation. Zhu et al. [46] introduced a graph contrastive representation learning model with adaptive augmentation by merging some priors for semantic aspects and topological of the graph. What's more, to add more noise to unimportant node feature to recognize the semantic information. You et al. [47] presented a GraphCL model of graph augmentations to combine various priors.

## 3. Preliminaries

### 3.1 Graph convolutional networks (GCNs)

Firstly, we apply an undirected graph $G = (V, E)$. $V$ is the set of nodes. $V = \{v_1,...v_M\}$. M represents the number of nodes. $E$ is the set of edges in which every element represents the two nodes linked. $E = \{e_1,...e_N\}$. N represents the number of edges. $M \in R^{M*D}$ represents the feature matrix. D is the dimensionality. $A \in R^{M*D}$ represents the adjacent matrix. We used the following formula to describe two nodes connected or not.

$$A_{ab} = A_{ba} = \begin{cases} 0, & \text{There are no links between a and b;} \\ 1, & \text{There are links between a and b;} \end{cases} \quad (1)$$

Finally, let's follow GCN[24] to describe a multi-layer Graph Convolutional Network.

$$H^{(l)} = \sigma(\dot{M}^{-\frac{1}{2}} \dot{A} \dot{M}^{-\frac{1}{2}} H^{(l-1)} W^{(l)}) \quad (2)$$

At this formula, $H^{(l)}$ is the matrix of activation functions in the layer of l, and $\dot{A}$ denotes the adjacent matrix of the undirected graph. $\sigma(\cdot)$ represents the activation function. $\dot{M}$ and $W^{(l)}$ represent the weight matrix.

### 3.2 Link prediction

We use the following formula to describe link prediction, which follows Zhu et al. [29].

$$M = \text{sigmoid}(F^i (F^i)^T) \quad (3)$$

Here, $F^i$ denotes the feature of nodes after learning by GCNs. M is the adjacent matrix. The specific steps of link prediction are as follows:

---

**Input:**
- $A_{adj}$: An adjacent matrix after deleting some links.
- $A_{old}$: An array that represents the row and column number of deleted links.
- $A_{original}$: An original adjacent matrix.

**Procedure Begin:**
1. Do the following operation.
   Transform $A_{original}$ to $A_{newp}$.
   Do for i ∈ { 0, 1 ... len($A_{newp}$)-1 }
      Perform following operations
      Do for j ∈ { 0, 1 ... i+1 }.
         Judge $A_{adj}$[i][j] is 1 or not.
         If $A_{adj}$[i][j] is 1, then save [[i, j], 1] to a new array, $A_{new}$.
         Else make the next loop.
   Output the new array, $A_{new}$.
   **End**
2. Do the following operation.
   To calculate the length of $A_{new}$, then get the result: num.
   Transform $A_{original}$ to $A_{newn}$.
   Do for i ∈ { 0, 1 ... len($A_{newn}$)-1 }

Perform following operations

Do for j ∈ { 0, 1 … i+1}.

Judge $A_{adj}[i][j]$ is 0 or not.

If $A_{adj}[i][j]$ is 0, then save [[i, j], -1] to a new array, $A_{newa}$.

Else make the next loop.

Truncate the first num element of $A_{newa}$, then output the result: $A_{newb}$.

**End**

3. Insert $A_{newb}$ to the end of $A_{new}$.
4. To calculate the length of $A_{old}$. The result is $L_{old}$.

Transform $A_{original}$ to the format of the array. The result is $A_{newa}$.

Do the following operation.

Do for i ∈ { 0, 1 …$L_{old}$-1}.

To get the random row number of $A_{newa}$. The result is row.

To get the random column number of $A_{newa}$. The result is col.

When $A_{newa}$[row][col] is 0 and [row, col] not in $A_{newb}$ and [col, row] not in $A_{newb}$, then save [row, col] to a new array. The result is $A_{newc}$.

Sort and output $A_{newc}$.

**End**

5. To loop $A_{old}$ and $A_{newc}$, then save the row and column number to a new array, respectively. The result is $A_{newd}$ and $A_{newe}$.
6. Input $A_{newd}$ and $A_{newe}$ to the update function, then get the result $R_{best}$.

**End**

**Output:** The result after link prediction: $R_{best}$.

Please notice that in Step 6 for **Algorithm1**, several points need to be described.

1) When $A_{newb}$ transforms to $A_{newp}$ and $A_{newn}$, it uses the csr_matrix function, which is the embedding method of scipy.

2) In step 4, before getting the random row and column number of $A_{newa}$, the total rows and columns of $A_{newa}$ are calculated. The shaping method is used to get them.

3) After executing our RRLFSOR strategy, then we get the $A_{adj}$ and $A_{old}$. In short, $A_{adj}$ and $A_{old}$ are the outputs of our RRLFSOR strategy.

## 4. Methods

Before introducing our strategy, let's describe the system architecture. It is presented in Fig. 1.

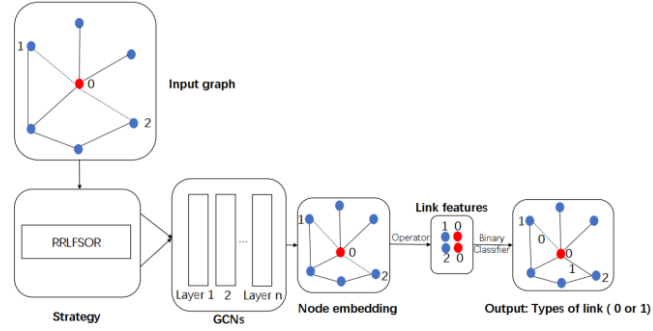

**Fig. 1.** System architecture

There are four parts to the system architecture. The first part is original graphs data. The actual graphs data is from the datasets. The second part is the strategy we proposed. The strategy is randomly removing links with a fixed step at one region (**RRLFSOR**). The third part is the GCNs. This part represents different variants of GCNs. To verify the superiority and efficiency of our strategy, we use two efficient and representative GCN models. For example, GCN and L-GCN. The last part is link prediction, which is a self-supervised learning task. The link prediction consists of node embedding, link features, and the output. The operator is the Hadamard. There are two types of results. The first one is 0. The second one is 1. If we define the operated adjacency matrix as $\dot{A}_{RRLFSOR}$, the original adjacency matrix is $\dot{A}$, then the result adjacency matrix $\dot{A}_{sparse}$ as following

$$\dot{A} = \dot{A}_{RRLFSOR} + \dot{A}_{sparse} \quad (4)$$

$$\dot{A}_{RRLFSOR} = P \otimes D_{step} \quad (5)$$

where $P$ is the percent of deleting links and $D_{step}$ is the step number. $\dot{A}_{RRLFSOR}$ is generated by RRLFSOR, respectively. In the following sections, we will describe RRLFSOR.

**4.1 Randomly remove links with a fixed step at one region (RRLFSOR)**

Based on the analysis of section 3.1, if there is a connection between two nodes, it is represented by 1; otherwise, use 0. If we want to delete one link, we have to change 1 to 0. We propose the first self-supervised learning strategy named randomly removing links with a fixed step at one region (RRLFSOR). Fig. 2 is the operation diagram of RRLFSOR.

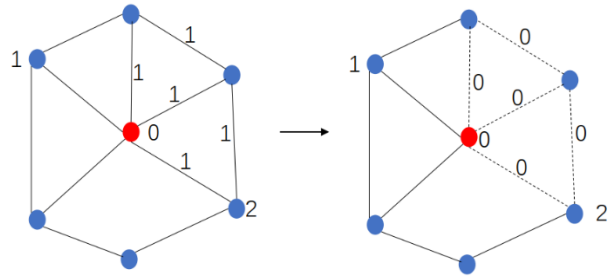

**Fig. 2.** Operation diagram of RRLFSOR

The first step of the RRLFSOR is told the percentage of deletions and the number of steps. Then, we randomly find one line to delete. The step means how many deletions are in one line. The specific steps of RRLFSOR are as follows:

---

**Input:**
- $A_{adj}$: The original adjacent matrix.
- P: The percentage of deletions.
- $D_{step}$: The number of steps.

**Procedure Begin:**
1. Calculate the length of $A_{adj}$. The result is $T_{adj}$.
2. Calculate the total number of deletions by P and $T_{adj}$. The result is $T_{del}$.
3. Calculate the row number of $A_{adj}$. The result is $T_{row}$.
4. Get the random row number from $T_{row}$. The result is $T_{rand}$.
5. Define an array that is saved the row and column number of one deletion. The result is $A_{del}$. Its length is $T_{del}$.
6. Do the following operation.
    1. Do while $T_{del} < T_{del}$.
        1. Get all index of $A_{adj}$ in row $T_{row}$. The result is $A_{index}$.
        2. Calculate the length of $A_{index}$. The result is $T_{index}$.
        3. Judge $T_{index}$ is more significant than 1. If it is, do the following operations. Otherwise, give up the procedure.
            Do for index ∈ {0, 1 …$D_{step}$ -1}
            1. Get the sub-column from $A_{index}$. The result is $C_{sub}$.
            2. Judge $A_{adj}[T_{rand}][ C_{sub}]$ or $A_{adj}[C_{sub}][ T_{rand}]$ is zero or not. If it is zero, give up current loop and execute next loop.
            3. Set $A_{adj}[T_{rand}][ C_{sub}]$ and $A_{adj}[C_{sub}][ T_{rand}]$ as zero.
            4. Save $T_{rand}$ and $C_{sub}$ to $A_{del}$.
        4. Add one to $T_{row}$. The purpose of this operation is to execute adjacent rows.
    2. Sort $A_{del}$.

**End**

**Output:** $A_{del}$ and $A_{adj}$

---

Please notice for the **Algorithm2**. Several points need to be described.

1) In step 1, we use the nnz of the adjacent matrix and divide two to get its length.

2) In step 2, $T_{del}$ may be a decimal. So we transform it by int function compulsorily.

3) We judge $T_{index}$ is more significant than one or not because we want to prevent isolated nodes.

## 4.2 Why does RRLFSOR improve the performance significantly?

**Prevent over-fitting.** Like the DropEdge skill, RRLFSOR generates different deformations of the input data also and can be regarded as one new data augmentation technique for graphs. DropEdge only randomly removes certain percent links of the input data and ignores the neighbours. The theory of GCNs is by gathering neighbours' information for each node. So we should consider the neighbours' information instead of randomly removing certain percent links only. Sometimes randomly removing certain percent links generate too sparse a subgraph and ignore the influence of neighbours' node. RRLFSOR not only consider randomly removing certain percent links but also consider neighbours' nodes. RRLFSOR is the variant of DropEdge, it generates one random subset aggregation instead of the full aggregation during GCNs training. In all, RRLFSOR is one augmentation skill for GCNs training, similar to image augmentation skill (random erasing) that are helpful to tackle over-fitting in training GCNs. Like DropEdge, we support an intuitive understanding of why RRLFSOR is valid.

## 5. Experiments and analysis

### 5.1 Datasets

To verify the effectiveness and superiority, we use three popular and representative datasets to test our strategy. Table 1 is the description of the three datasets.

**Table 1** Summary of the datasets

| Dataset | Features Number | Nodes Number | Edges Number | Classes |
|---|---|---|---|---|
| Cora | 1433 | 2708 | 5429 | 7 |
| PubMed | 500 | 19717 | 44338 | 3 |
| Citeseer | 3327 | 3327 | 4732 | 6 |

### 5.2 Experimental setting

Due to the strategy is to remove links with a fixed step in each row, we need to calculate the maximum number of links in one row. Table 2 describes the maximum number of links in one row of each dataset.

**Table 2** The maximum number of links in one row of each dataset

| Dataset | Cora | PubMed | Citeseer |
|---|---|---|---|
| The max links number in one row | 168 | 99 | 125 |

To set a reasonable step size of deletions, we should describe the times of different step sizes in the dataset. Table 3 describes the quantitative relationship.

**Table 3** The times of different step sizes appear in the dataset

| Step | Cora | PubMed | Citeseer |
|---|---|---|---|
| 1 | 485 | **9094** | **1352** |
| 2 | **583** | 3357 | 805 |
| 3 | 553 | 1584 | 444 |
| 4 | 389 | 914 | 241 |
| 5 | 281 | 642 | 142 |
| 6 | 131 | 491 | 114 |
| 7 | 82 | 422 | 60 |

The size of hidden units of [9] is 16. Unlike the setting of the hidden unit size of [9], the size of hidden units in our L-GCN experiments is 270. When the size of hidden units is set at 270, the accuracy is improved by 0.5% than the original value. In the experiments of GCN+RRLFSOR, we set the hidden units as 270.

We set the learning rate as 0.001. The layer number is 2. The weight decay is 5e-4. Adam is chosen as the optimizer. The epoch is 5000, which is followed by [29]. We obtain the best results in 10 experiments.

### 5.3 Comparison of the strategy and Results analysis

We apply our RRLFSOR strategy to GCN. In Cora and Citeseer dataset, we set the percentage of deletions as 10%，20%, 30%, 40%,50% and 60% respectively. Table 4 describes the results. Bold font indicates the maximum and minimum improvements.

Table 4 The accuracy of GCN under the RRLFSOR strategy with different steps and percentages of deletions. (Unit: %)

| GCN | Cora | | | | | | Citeseer | | | | | |
|---|---|---|---|---|---|---|---|---|---|---|---|---|
| Without | 83.80 | | | | | | 70.30 | | | | | |
| Percentage of deletions | 10 | 20 | 30 | 40 | 50 | 60 | 10 | 20 | 30 | 40 | 50 | 60 |
| 1 | **94.12** | 91.15 | 90.48 | 89.28 | **85.69** | 80.66 | **91.64** | 90.25 | 90.48 | 90.48 | 85.68 | 84.78 |
| 2 | 81.83 | 81.08 | 82.12 | 80.94 | 78.32 | 72.98 | 90.35 | 90.24 | 90.46 | 90.48 | 85.69 | 84.79 |
| 3 | 82.30 | 78.20 | 79.80 | 75.57 | 78.15 | 74.69 | 90.25 | 90.24 | 90.47 | 90.48 | 85.68 | 84.77 |
| 4 | 83.46 | 78.28 | 80.11 | 77.90 | 76.43 | 71.36 | 90.25 | 90.25 | 90.46 | 90.46 | 85.66 | **84.75** |
| 5 | 82.39 | 80.66 | 77.81 | 79.59 | 77.61 | 72.68 | 90.25 | 90.25 | 90.48 | 90.48 | 85.68 | 84.76 |
| 6 | 89.33 | 83.47 | 82.53 | 79.45 | 81.32 | 73.11 | 90.24 | 90.24 | 90.47 | 90.47 | 85.67 | 84.78 |

According to Table 4, it's not difficult to find: If we don't apply RRLFSOR to GCN, the accuracy is 83.80% in the Cora dataset. If we don't apply RRLFSOR to GCN, the accuracy is 70.30% in the Citeseer dataset. When the dataset is Citeseer, the step is 1, and the percentage of deletions is 10%, GCN+RRLFSOR gets the highest accuracy, 91.64%. The highest accuracy improves 21.34% than GCN. It could verify the efficient of RRLFSOR.

To better describe the process of the maximum accuracy of Table 4, we depict the loss value, which is illustrated in Fig 3.

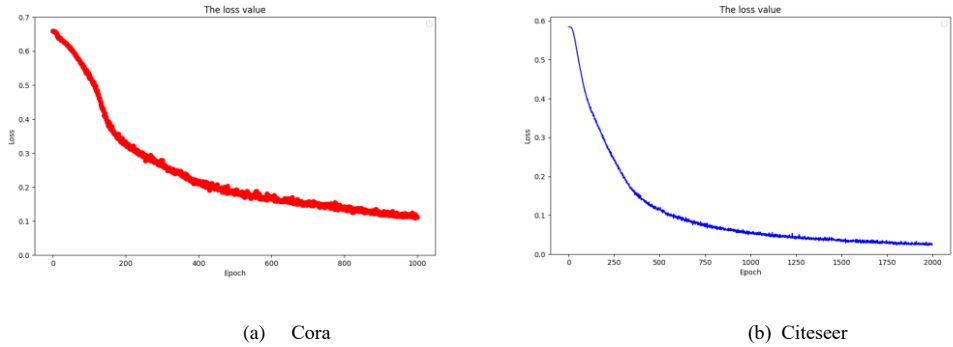

(a) Cora         (b) Citeseer

**Fig 3.** The loss value when the accuracy is the highest value of Table 4

According to Fig 3, it's not difficult to find: **1)** As the epoch keeps increasing, the loss keeps decreasing. **2)** When the epoch is 1000, the accuracy is lowest in the Cora dataset. **3)** When the epoch is 2000, the accuracy is lowest in the Citeseer dataset. The loss is decreased further which indicates the significance of RRLFSOR in alleviating over-smoothing.

Through experiments, it is found that when the percentage of deletions is 10%, the accuracy achieves the maximum value in the PubMed dataset. So we set the percentage of deletions as 10%, and the step is 1, 2, 3, 4, 5, and 6. We calculate the accuracy of GCN+RRLFSOR under the PubMed dataset. Table 5 describes the results. Bold font indicates the maximum value.

Table 5 The accuracy of GCN+RRLFSOR under PubMed dataset (Unit: %)

| GCN | PubMed |
|---|---|
| Without | 79.00 |

| Step | Acc | Epoch |
|---|---|---|
| 1 | 96.48 | 1000 |
| 2 | 97.12 | 6000 |
| 3 | **97.33** | 7500 |
| 4 | 96.89 | 5000 |
| 5 | 96.90 | 5000 |
| 6 | 96.83 | 5000 |

According to Table 5, it's not difficult to find: When the step is 3 and the epoch is 7500, we get the highest accuracy. The value is 97.33%. When the step is less than 4, as the epoch keeps increasing, the accuracy increases. When the step is more than 3, the accuracy decreases.

We apply our RRLFSOR strategy to L-GCN. In the two datasets, we set the percentage of deletions as 10%，20%, 30%, 40%,50%, 60% respectively. Table 6 describes the results. Bold font indicates the maximum and minimum growth. "/" indicates that the datasets do not provide the results.

Table 6 The accuracy of L-GCN under the RRLFSOR strategy with different steps and percentages of deletions. (Unit: %)

| L-GCN | Cora | | | | | | Citeseer | | | | | |
|---|---|---|---|---|---|---|---|---|---|---|---|---|
| Without | 85.20 | | | | | | / | | | | | |
| Percentage of deletions | 10 | 20 | 30 | 40 | 50 | 60 | 10 | 20 | 30 | 40 | 50 | 60 |
| 1 | **90.51** | 87.57 | **86.73** | 84.75 | 82.81 | 80.37 | **87.14** | 84.79 | 80.93 | 78.39 | 77.62 | 77.10 |
| 2 | 82.88 | 80.70 | 81.72 | 79.34 | 75.37 | 74.45 | 81.95 | 79.58 | 76.14 | 80.29 | 79.33 | 76.32 |
| 3 | 80.38 | 79.83 | 80.64 | 78.59 | 73.96 | 74.62 | 81.25 | 79.67 | 74.47 | 79.20 | 79.21 | 76.19 |
| 4 | 80.77 | 82.59 | 81.79 | 79.08 | 74.60 | 74.61 | 77.71 | 77.33 | 77.66 | 79.16 | 78.58 | 75.29 |
| 5 | 76.20 | 80.32 | 81.32 | 74.48 | 78.12 | 77.90 | **72.47** | 80.67 | 78.43 | 80.44 | 78.19 | 75.17 |
| 6 | 82.61 | 82.26 | 84.05 | 80.11 | 81.34 | 78.67 | 80.90 | 80.13 | 73.36 | 81.40 | 79.90 | 76.88 |

According to Table 6, it's not difficult to find: **1)** The highest accuracy is 90.51% in the Cora dataset when the step is 1, and the percentage of deletions is 10%. The result improves by 5.31% than L-GCN in Cora dataset. **2)** When the step is one, and the percentage of deletions is 30%, GCN+RRLFSOR gets the minimum increase in the Cora dataset. The result is 86.73%, which improves 1.53% than L-GCN. **3)** In the Cora dataset, regardless of the percentage of deletions, as long as the step is 1, the accuracy is the highest in Table 6. **4）** When the step is 1. The percentage of deletions is 10%, and the hidden size is 270. GCN+RRLFSOR gets the highest increase in the Cora and Citeseer datasets.

Through experiments, it is found that when the percentage of deletions is 10%, the accuracy achieves the maximum value in the PubMed dataset. So the percentage is 10%, and the step is 1, 2, 3, 4, 5, and 6. We calculate the accuracy of L-GCN+RRLFSOR under the PubMed dataset. Table 7 describes the results. Bold font indicates the maximum value.

Table 7 The accuracy of L-GCN+RRLFSOR under PubMed dataset (Unit: %)

| L-GCN | PubMed | |
|---|---|---|
| Without | 88.80 | |
| Step | Acc | Epoch |
| 1 | **97.19** | 6000 |
| 2 | 94.94 | 5000 |
| 3 | 95.19 | 5000 |
| 4 | 95.88 | 5000 |
| 5 | 96.04 | 5000 |
| 6 | 92.31 | 5000 |

According to Table 7, it's not difficult to find: **1)** When the epoch increases from 2 to 5, the accuracy increases in turn. **2**）The maximum accuracy value is 97.19% in the PubMed dataset, which improves 8.39% than L-GCN. The minimum accuracy value is 92.31% in the PubMed dataset, which improves 3.51% than L-GCN.

The performances of GCN and L-GCN combined with the RRLFSOR strategy on the two datasets are described in Table 8. The highest accuracy is highlighted in red for each dataset. "/" means not reported.

Table 8 The accuracy of two datasets. (Unit: %)

| Model | Cora | PubMed | Citeseer |
|---|---|---|---|
| Geom-GCN[5] | 85.65 | 90.49 | 79.41 |
| L-GCN[9] | 85.20 | 88.80 | / |
| GCN[26] | 83.80 | 79.00 | 70.30 |
| GAT[27] | 83.70 | 79.30 | 73.20 |
| GCN+SSL[29] | 84.53 | 82.09 | 73.57 |
| GAT+SSL[29] | 84.31 | 79.67 | 73.45 |
| E-GCN[30] | 84.60 | 80.70 | 74.80 |
| N-GCN[19] | 83.00 | 79.50 | 72.20 |
| Cross-GCN[32] | 81.40 | 80.60 | 71.80 |
| GCN+RRLFSOR(Ours) | 94.12 | 97.33 | 91.64 |
| L-GCN+RRLFSOR(Ours) | 90.51 | 97.19 | 87.14 |

From Table 8, we can find that: **1)** GCN+RRLFSOR obtain the highest accuracy in the Cora dataset. The result is 94.12%. **2)** GCN+RRLFSOR obtain the highest accuracy in the Citeseer dataset. The result is 91.64%. **3)** GCN+RRLFSOR obtain the highest accuracy in the PubMed dataset. The result is 97.33%. **4)** GCN+RRLFSOR improve 10.32% than GCN in Cora dataset. **5)** GCN+RRLFSOR improve 21.34% than GCN in Citeseer dataset. **6)** GCN+RRLFSOR improve 18.33% than GCN in PubMed dataset. **7)** GCN+RRLFSOR improves 19.84% than Cross-GCN in Citeseer dataset. **8)** L-GCN+RRLFSOR improve 8.39% than L-GCN in PubMed dataset. **9)** L-GCN+RRLFSOR improves 16.59% than Geom-GCN in PubMed dataset.

### 5.4  Ablation Study

We configure different numbers of hidden units to understand their influence on it. Table 9 is the parameter configuration table in this experiment. "/" means that this parameter is not needed in this model.

Table 9 The epoch configuration in the experiments

| Model | Step | The percentage of deletions | Dataset | Epoch Number |
|---|---|---|---|---|
| GCN | / | / | Cora | 200 |
| L-GCN | / | / | Cora | 160 |
| GCN+RRLFSOR | 1 | 10% | Cora | 1000 |
| L-GCN+RRLFSOR | 1 | 10% | Cora | 5000 |

We set the hidden units as 8, 16, 32, 64, 128, 256, 270 and 512. The accuracy is reported in Table 10. Bold font indicates the maximum accuracy in current model.

Table 10 The accuracy of models under different hidden units. (Unit: %)

| Models | The size of hidden units | | | | | | | |
|---|---|---|---|---|---|---|---|---|
| | **8** | 8+8 | 16+16 | **32+32** | 64+64 | 128+128 | **256+14** | 256+256 |
| GCN | 80.30 | 82.90 | **83.80** | 83.10 | 82.50 | 83.10 | 83.10 | 83.30 |
| L-GCN | **85.30** | 84.60 | 84.30 | 84.60 | 84.80 | 84.70 | 85.20 | 84.70 |
| **GCN+RRLFSOR** | 74.55 | 88.01 | 92.20 | 93.26 | 93.85 | 93.82 | **94.12** | 93.86 |
| **L-GCN+RRLFSOR** | 81.27 | 88.30 | 89.60 | 90.41 | 89.16 | 88.35 | **90.51** | 87.94 |

According to Table 10, it's not difficult to find: **1)** The highest accuracy of GCN is 83.80% when the size of hidden units is 32. **2)** When the hidden units are 8, GCN obtains minimal accuracy in the Cora dataset. **3)** GCN+RRLFSOR gets the highest accuracy when the hidden unit is 270. The result is 94.12%. **4)** The highest accuracy of GCN+RRLFSOR improves 10.32% than GCN. **5)** The highest accuracy of L-GCN+RRLFSOR improves 5.21% than L-GCN when the hidden units are 270.

To better describe the process of the accuracy of Table 13, we depict the loss value of them when the hidden units are 8, 64 and 270 which is described in Figure 4. What's more, when the hidden size takes 8, we call it Hidden8. When the hidden size takes 64, we call it Hidden64. When the hidden size takes 270, we call it Hidden270.

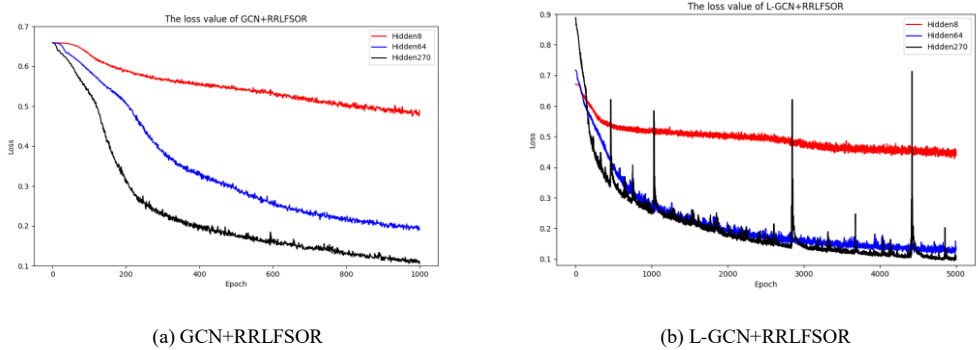

(a) GCN+RRLFSOR    (b) L-GCN+RRLFSOR

Fig 4 Different loss under different units

According to Fig 4, it's not difficult to find: **1)** As epochs continue to increase, the loss value as a whole continues to decrease. **2)** In GCN+RRLFSOR, the loss value of Hidden8 is bigger than Hidden64 and Hidden270. **3)** When the Hidden is 270, the loss value is smaller which could also verify it's efficient in alleviating over-smoothing.

**5.5 Result summary**

In short, we summarize the results of RRLFSOR as follows.

Firstly, whether the step is equal to 1 or greater than 1, adding RRLFSOR to GCNs significantly increases accuracy. We use GCN and L-GCN as examples for verification.

Secondly, in most cases, when the step is set to 1, the accuracy is the highest.

Thirdly, in the PubMed dataset, when RRLFSOR is applied to L-GCN, the accuracy increases by up to 8.39%.

What's more, when RRLFSOR is applied to GCN, the accuracy increases by up to 10.32% in the Cora dataset.

Specifically, RRLFSOR can be regarded as the generation of DropEdge[25] and Dropout[49]. Dropout only alleviated over-fitting but cannot solve the over-smoothing because it never changed the adjacency matrix. DropEdge was proved as complementary with Dropout. That is to say, the improvement by DropEdge is more prominent than Dropout. However, when they are adopted concurrently, the validation loss is decreased further. Due to the space limit, we will leave the experiment on our strategy with Dropout for future exploration. We assume that our strategy and Dropout are complementary to each other also.

Last but not least, in the Citeseer dataset, when RRLFSOR is applied to GCN, the accuracy increases by up to 23.31%.

## 6. Conclusions

We proposed one strategy to address GNN and GCNs requiring any labelled data. To verify the superiority and effectiveness of our strategy, we performed link prediction experiments using three public citation network datasets on two efficient and representative GCN models. Extensive experiments show that our strategy achieves remarkable performance improvement on GCNs.

In future works, the time complexity of RRLFSOR should be improved. Because when the total edges are more than 100,000, it will waste more time finding and deleting suitable links which require high memory costs. What's more, we will explore the improvement by our strategy and Dropout.

**Acknowledgements**

*We thank anonymous reviewers for valuable comments that help improve the paper during revision. We would rather thank Yuning You for providing practical details on [9] and [18].*

This work was supported in part by the National Natural Science Foundation of China (Grant No.61863005, No.62163007), the Science and Technology Foundation of Guizhou Province (Grant No. QKHZC[2019]2814, No. [2020]4Y056, No. PTRC[2020]6007, No. [2021]439, No. QKHPT-JXCX[2021]001), the Construction Project of the Integrated Tackling Platform for Colleges and Universities in Guizhou Province (Grant No. KY[2020]005), the Experimental Technology and Development Project of Zhejiang Normal University(Grant No.SJ202123), and the Digital Reform Project of Zhejiang Normal University(Grant No. [2021]05).